\documentclass[review]{fcs}
\usepackage{multirow}
\usepackage{booktabs}
\usepackage{tcolorbox}
\usepackage{diagbox}
\usepackage{listings}
\usepackage[edges]{forest}
\usepackage[numbers]{natbib}

\definecolor{hidden-draw}{RGB}{20,68,106}
\definecolor{hidden-pink}{RGB}{255,245,247}

\title{Large Language Model for Table Processing: A Survey}
\author[1]{Weizheng Lu}
\author*[1]{Jing Zhang}
\author[1,2]{Ju Fan}
\author[3]{Zihao Fu}
\author*[1,2]{Yueguo Chen}
\author[1,2]{Xiaoyong Du}
\address[1]{School of Information, Renmin University of China, Beijing 100872, China}
\address[2]{Key Laboratory of Data Engineering and Knowledge Engineering, Beijing 100872, China}
\address[3]{WPS Office, Kingsoft Co. Zhuhai 519080, China}
\fcssetup{
  received       = {month dd, yyyy},
  accepted       = {month dd, yyyy},
  corr-email     = {luweizheng,zhang-jing,chenyueguo@ruc.edu.cn},
}

\begin{abstract}
Tables, typically two-dimensional and structured to store large amounts of data, are essential in daily activities like database queries, spreadsheet manipulations, web table question answering, and image table information extraction. Automating these table-centric tasks with Large Language Models (LLMs) or Visual Language Models (VLMs) offers significant public benefits, garnering interest from academia and industry.
This survey provides a comprehensive overview of table-related tasks, examining both user scenarios and technical aspects. It covers traditional tasks like table question answering as well as emerging fields such as spreadsheet manipulation and table data analysis. 
We summarize the training techniques for LLMs and VLMs tailored for table processing. Additionally, we discuss prompt engineering, particularly the use of LLM-powered agents, for various table-related tasks. 
Finally, we highlight several challenges, including diverse user input when serving and slow thinking using chain-of-thought.
\end{abstract}

\keywords{Data Mining and Knowledge Discovery, Table Processing, Large Language Model}

\begin{document}

\section{Introduction}

In this data-driven era, a substantial volume of data is structured and stored in the form of tables~\cite{dong2022Table, badaro2023Transformers}. 
Everyday tasks involving tables, such as database queries, spreadsheet manipulations, question answering on web tables, and information extraction from image tables are common in our daily lives. 
While some of these tasks are tedious and error-prone, others require specialized skills and should be simplified for broader accessibility. 
The automation of table-related tasks provides substantial benefits to the general public, garnering significant interest from both academic and industrial sectors~\cite{badaro2023Transformers, fang2024Large, zhang2024Survey}.

Recently, large language models (LLMs) have demonstrated their effectiveness and versatility across diverse tasks, leading to significant advancements in natural language processing~\cite{zhao2023Survey}. This success has spurred researchers to investigate the application of LLMs to table-related tasks. However, the structure of tables differs from the plain text~\cite{raffel2020Exploring} typically used during LLM pre-training. 

\begin{itemize}[leftmargin=*]

\item \textbf{Structured Data} Tables are inherently structured, composed of rows and columns, each with its own schema that outlines the data's semantics and their interrelations. Humans can effortlessly interpret tables both vertically and horizontally, but LLMs, primarily trained with sequential text data, struggle with understanding the multidimensional aspects of tables.

\item \textbf{Complex Reasoning} Tasks in table processing often require numerical operations (like comparisons or aggregations), data preparation (such as column type annotation and missing value detection), and more sophisticated analyses (including feature engineering and visualization). These tasks demand intricate reasoning, the ability to decompose problems into multiple steps, and logical operations, thereby posing significant challenges to machine intelligence.

\item \textbf{Utilizing External Tools} In real-world scenarios, humans often depend on specialized tools such as Microsoft Excel, Python, or SQL for interacting with tables. For effective table processing, LLMs need to be adept at integrating and using these external tools.

\end{itemize}

Although many text-related tasks, such as those in STEM (Science, Technology, Engineering, and Mathematics) fields, require complex reasoning and external tools, table processing tasks are different due to the structural nature of tables and the user intent of querying knowledge from tables. For instance, LLMs need to understand table schemas, locate data within two-dimensional tables, and execute SQL queries to retrieve data.
The unique challenges presented by table processing tasks emphasize the need to tailor LLMs for these specific purposes.
Early research, such as TaBERT~\cite{yin2020tabert}, \\TaPas~\cite{herzig2020TaPas}, TURL~\cite{deng2020TURL}, and TaPEx~\cite{liu2022TAPEX}, adhere to the paradigm of pre-training or fine-tuning neural language models for tables. These methods adapt model architectures, including position embeddings, attention mechanisms, and learning objectives for pretraining tasks.
While these approaches yield good results, they are largely confined to specific table tasks like table question answering (table QA) and fact verification. Additionally, the BERT or BART models they utilize are not sufficiently large or versatile to handle a broader range of table tasks. 
Latest LLM-based approaches tackle table tasks in two primary ways: (1) curating table datasets and pre-train or fine-tune a table model~\cite{zhang2023TableLlama,li2024TableGPT, zheng2024Multimodal}; (2) prompting an LLM or building an LLM-powered agent by utilizing the LLM's strong reasoning ability to understand table data~\cite{sui2024Table, zhang2024ReAcTable, li2024SheetCopilot, hu2024InfiAgentDABench}.
These newer methods leverage LLM-specific technologies, such as instruction-tuning~\cite{wei2022Finetuned}, in-context learning~\cite{brown2020Language}, chain-of-thought reasoning~\cite{wei2022Chainofthought}, and autonomous agents~\cite{wang2023AgentSurvey}, showcasing a more versatile and comprehensive approach to table processing.

\tikzstyle{my-box}=[
	rectangle,
	draw=hidden-draw,
	rounded corners,
	text opacity=1,
	minimum height=1.5em,
	minimum width=5em,
	inner sep=2pt,
	align=center,
	fill opacity=.5,
	line width=0.8pt,
]
\tikzstyle{leaf}=[my-box, minimum height=1.5em,
	fill=hidden-pink!80, text=black, align=left,font=\normalsize,
	inner xsep=2pt,
	inner ysep=4pt,
	line width=0.8pt,
]
\begin{figure*}[ht]
	\centering
	\resizebox{\textwidth}{!}{
		\begin{forest}
			forked edges,
			for tree={
				grow=east,
				reversed=true,
				anchor=base west,
				parent anchor=east,
				child anchor=west,
				base=center,
				font=\large,
				rectangle,
				draw=hidden-draw,
				rounded corners,
				align=left,
				text centered,
				minimum width=4em,
				edge+={darkgray, line width=1pt},
				s sep=3pt,
				inner xsep=2pt,
				inner ysep=3pt,
				line width=0.8pt,
				ver/.style={rotate=90, child anchor=north, parent anchor=south, anchor=center},
			},
			where level=1{text width=13em,font=\normalsize,}{},
			where level=2{text width=9em,font=\normalsize,}{},
			where level=3{text width=28em,font=\normalsize,}{},
			where level=4{text width=18em,font=\normalsize,}{},
			where level=5{text width=18em,font=\normalsize,}{},
			[
				\textbf{LLM for Table Processing}, ver
				[
					\textbf{Table Types and Table Tasks}  (\S \ref{sec:table_types_tasks}), fill=blue!10
					[
						\textbf{Table Definition}  (\S \ref{subsec:table_def}), fill=blue!10
						[
							\textbf{Spreadsheet, Web Table, Database, Document}, fill=blue!10
						]
					]
					[
						\textbf{Table Tasks} (\S \ref{subsec:table_tasks}), fill=blue!10
						[
							\textbf{Table QA, Spreadsheet manipulation, NL2SQL, Table extraction, etc.}, fill=blue!10
						]
					]
				]
				[
					\textbf{Table Data Representation}  (\S \ref{sec:representation}), fill=orange!10
					[
						\textbf{Text} (\S \ref{subsec:text_repr}),  fill=orange!10
					]
					[
						\textbf{Visual and Layout} (\S \ref{subsec:visual_repr}), fill=orange!10
					]
				]
				[
					\textbf{Table Training} (\S \ref{sec:train_method}), fill=green!10
					[
    					\textbf{LLM} (\S \ref{subsec:llm_training}), fill=green!10
                        [
							\textbf{Continue Pre-training, Instruction Tuning, etc.}, fill=green!10
						]
					]
                    [
    					\textbf{VLM} (\S \ref{subsec:vlm_training}), fill=green!10
                        [
							\textbf{Encoder Pretraining + Fine-tuning}, fill=green!10
						]
					]
				]
                [
					\textbf{Table Prompting} (\S \ref{sec:prompt}), fill=black!10
					[
    					\textbf{Planning} (\S \ref{subsec:planning}), fill=black!10
                        [
							\textbf{Complex Task Decomposition, Reflection \& Revision, etc.}, fill=black!10
						]
					]
                    [
    					\textbf{Action} (\S \ref{subsec:action}), fill=black!10
					]
				]
			]
		\end{forest}
	}
	\caption{Taxonomy of LLMs for table processing.}
	\label{fig:taxonomy_table_llm}
\end{figure*}
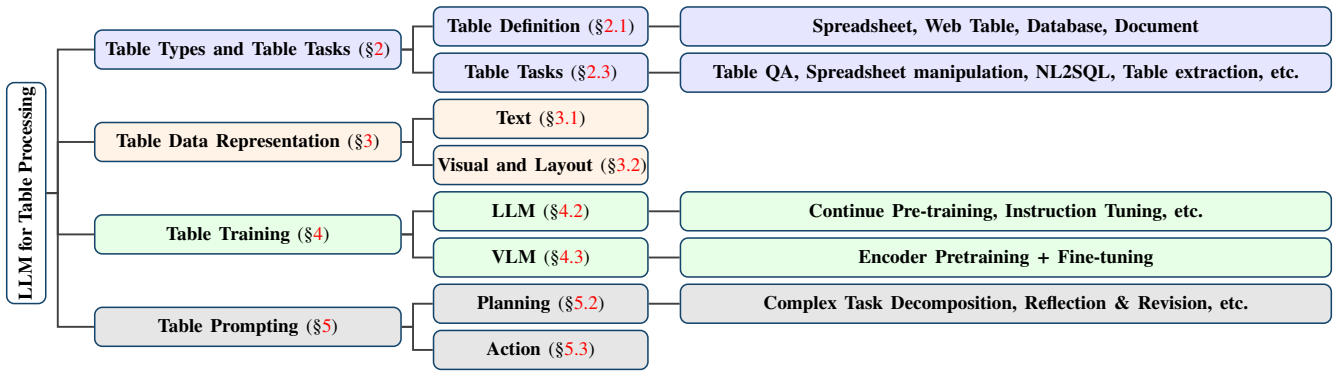

\textbf{Taxonomy} The goal of this survey is to offer a comprehensive review of technological advancements in LLM for table processing and to summarize current research directions.
As depicted in Figure~\ref{fig:taxonomy_table_llm}, we have categorized the literature into a taxonomy of four key categories: table types and table tasks, table data representation, table training, and table prompting.
These four categories cover distinct and interrelated research topics, offering a systematic and comprehensive review of LLM for table processing research.

\textbf{Contribution} The main contribution of this survey is its extensive coverage of a wide range of table tasks, including recently proposed spreadsheet manipulation and data analysis. 
We discuss table tasks not only from a technical perspective but also from the table data lifecycle and from the end-user's viewpoint.
We categorize methods based on the latest paradigms in LLM usage, focusing on instruct\-ion-tuning, data synthesis, chain-of-thought, ReAct, and LLM-powered agent approaches. 
We compile recent datasets, benchmarks, and training corpora.
We collect resources such as papers, code, and data\-sets, which can be accessed at our website~\cite{llm-table-survey}.

\textbf{Comparison with Related Surveys} 
Earlier surveys, such as those by Dong et al.~\cite{dong2022Table} and Badaro et al.~\cite{badaro2023Transformers}, primarily concentrate on pre-training or fine-tuning techniques using smaller models like BERT~\cite{herzig2020TaPas,yin2020tabert} or BART~\cite{liu2022TAPEX}. However, they do not address methods based on LLMs, particularly those involving prompting strategies and agent-based approaches. 
Additionally, some surveys are confined to limited table tasks. For instance, Jin et al.~\cite{jin2022Survey} focus solely on table QA. Zhang et al.~\cite{zhang2024Survey} focus on table reasoning, overlooking tasks such as spreadsheet manipulation.
Fang et al.~\cite{fang2024Large} review research on table data prediction, generation, and understanding; however, the discussion on table processing lacks depth. 
Qin et al.~\cite{qin2022Survey} and Hong et al.~\cite{hong2024NextGeneration} concentrate on natural language to SQL (NL2SQL), overlooking spreadsheet manipulation and data analysis tasks.
\section{Table Types and Table Tasks }
\label{sec:table_types_tasks}
 
Tables are prevalent data structures that organize and manipulate knowledge and information in almost every domain. We briefly summarize table formats, table tasks and table data lifecycle.

\subsection{Table Definition}
\label{subsec:table_def}

\begin{figure*}[ht]
    \centering
    \includegraphics[width=1\linewidth]{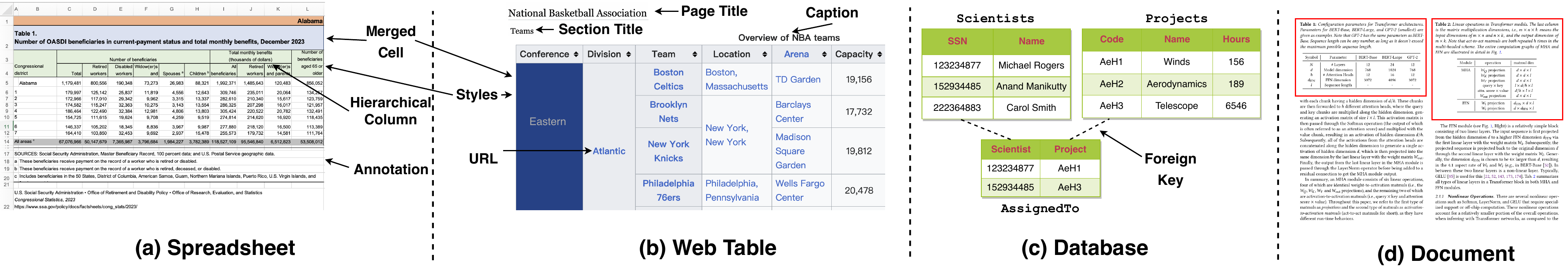}
    \caption{Four types of tables: Spreadsheet, Web Table, Database and Document.}
    \label{fig:data_format}
\end{figure*}

This paper mainly focuses on (semi-)structured tables, which are grids of cells arranged in rows and columns~\cite{zhang2020Web}.
Every column in a table signifies a specific attribute, each having its own data type (e.g., numeric, string, or date). Each row forms a record populated with diverse attribute values.
A cell is the intersection of a row and a column.
Cells are basic units to store text, numerical values, formulas, etc~\cite{dong2022Table}.
The two-dimensional structure endows the data with a schema, which includes column names, data types, constraints, and relationships with other tables. 
Tables are mainly stored and presented in the following formats:

\begin{itemize}[leftmargin=*]
    \item \textbf{Spreadsheet} (SS): Spreadsheets are used by one-tenth of the global population~\cite{rahman2020Benchmarking}, with Google Sheets and Microsoft Excel being the two most popular systems. Spreadsheet systems allow users to format tables by customizing styles (such as font or color) to display or highlight certain data. Spreadsheet tables often feature irregular layouts with merged cells, hierarchical columns, and annotations, as shown in Fig.~\ref{fig:data_format} (a). The irregular layout is designed to improve human understanding of the table data, but it makes machine parsing difficult. Spreadsheet systems provide features like sorting, filtering, formulas, data visualization, or table programming (e.g., Visual Basic for Applications, VBA) for automated or semi-automated data analysis. Typical applications of spreadsheets include: teachers in schools recording grades, human-resource departments recording employee information, and sales personnel tracking sales data, among others.
    \item \textbf{Web Table} (WT): The Web hosts a multitude of tables, created for diverse purposes and containing valuable information~\cite{Ritze2017t2dv2, bhagavatula2015TabEL, zhang2020Web}. These tables exist in various formats, from HTML to markdown, JSON, and XML. As web tables are embedded in web pages, they have much contextual information, such as the page’s title, the surrounding text, and so on. Web tables can also have various styles and embed URLs. A well-known web table example is the table on a \\Wikipedia page shown in Fig.~\ref{fig:data_format} (b). Web tables contain substantial factual knowledge, and the community has been working to extract and structure these tables.
    For example, the WDC (Web Data Commons) Web Table Corpus project converts billions of web pages from the Common Crawl corpus into structured tables~\cite{zhang2020Web}, and Bhagavatula et al.~\cite{bhagavatula2015TabEL} extract millions of tables from Wikipedia. The extraction process involves tasks like entity linking or relation extraction, and the extracted tables can be utilized for question answering.
    \item \textbf{Database} (DB):
    Tables in relational databases are highly structured, i.e., each database table is explicitly defined with a schema at the creation time. Users should use SQL to interact with database tables. Database systems are categorized into online transaction processing (OLTP) and online analytic processing (OLAP). OLTP systems frequently utilize foreign keys to establish relationships between tables, as illustrated in Figure \ref{fig:data_format} (c). Tables in OLAP systems or data warehouses, often comprising many columns, are called wide tables. Wide tables can minimize the need for complex joins. Database systems are scalable and robust and are employed in enterprises requiring data accuracy and integrity.
    Companies typically employ an IT team proficient in programming to manage these database systems.
    \item \textbf{Document} (DOC) :
    There is another category of tables embedded in various document formats, such as images ({\it .png}, {\it .jpg}, etc.), PDFs ({\it .pdf}), or Microsoft Word files ({\it .docx}). 
    Common examples of documents with tables include purchase orders, financial reports, sales contracts, receipts, academic papers (Fig.~\ref{fig:data_format} (d)), and numerous other types.
    Users want to extract tables from these documents, structure them, and convert them into table-native formats (spreadsheet or HTML). These tables are often surrounded by text, necessitating identifying their location before extracting their content, and the layout of these tables may be irregular. Furthermore, unlike ordinary images or plain text documents, document-embedded tables rely on an exact two-dimensional coordinate system. Any misalignment in the rows and columns can significantly impair the understanding of the information presented.
\end{itemize}

Since these four types of tables are tailored to different user scenarios and address diverse problems, integrating artificial intelligence (AI) models with these four types of tables varies accordingly. Spreadsheet systems aim to copilot users by automating various manipulation operations. Web tables can be used for table QA. Databases now widely incorporate NL2SQL technologies, aiding human engineers in data engineering and analytical tasks. Tables within documents need to be identified, structured, and transformed into table-native formats.


\subsection{Differences Between Table and Text}
Many AI methodologies migrate text modeling techniques to tables. Therefore, we should consider the differences between tables and text. 
Li et al.~\cite{li2024TableGPT} outlines the main distinctions between them. Texts are (1) one-directional; (2) typically read from left to right; and (3) the swapping of two tokens usually alters the sentence's meaning. 
On the other hand, tables are (1) two-dimensional, requiring both horizontal and vertical reading; (2) their understanding heavily relies on schemas or header names; and (3) some of them remain unaffected by row and column permutations.

\subsection{Table Tasks}
\label{subsec:table_tasks}

\begin{table*}[ht]
\centering
\caption{Table tasks, input types, descriptions (related work), and representative datasets. In the Table Type column, the abbreviations are as follows: WT for Web Table, SS for Spreadsheet, DB for Database, and DOC for Document.}
\label{tab:table_tasks}
\begin{tabular}{cccc}
\toprule
Task Name                                                            & Table Type & Description (related work)                                                                                                         & Example Dataset   \\ \hline
Table QA                                                             & WT         & Answer a NL question given a table (~\cite{cheng2023Binding, wang2024chainoftable})                                                                                              & WikiTableQuestion~\cite{pasupat2015wikitq} \\ \hline
\begin{tabular}[c]{@{}c@{}}Table \\ fact verification\end{tabular}   & WT         & Verifying facts given a table (~\cite{wang2024chainoftable, ye2023Large})                                                                                                   & TabFact~\cite{chen2019tabfact}           \\ \hline
Table-to-text                                                        & WT         & Produce a NL question given a table (~\cite{zhang2023TableLlama})                                                                                             & ToTTo~\cite{parikh2020ToTTo}             \\ \hline
Data cleaning                                                        & WT/SS/DB   & Correct errors of table data (~\cite{qian2024UniDM,ahmad2023RetClean})                                                                                                    & -                 \\ \hline
\begin{tabular}[c]{@{}c@{}}Column/Row/Cell\\ population\end{tabular} & WT/SS/DB   & Populate possible column/row/cell for a table (~\cite{zhang2023TableLlama,li2024TableGPT})                                                                             & TURL~\cite{deng2020TURL}              \\ \hline
Entity linking                                                       & WT         & Link the selected entity to the knowledge base (~\cite{zhang2023TableLlama,li2024TableGPT})                                                                                  & TURL~\cite{deng2020TURL}              \\ \hline
\begin{tabular}[c]{@{}c@{}}Column type \\ annotation\end{tabular}    & WT         & Choose types for the column in the table (~\cite{zhang2023TableLlama,li2024TableGPT})                                                                            & TURL~\cite{deng2020TURL}              \\ \hline
\begin{tabular}[c]{@{}c@{}}Spreadsheet \\ manipulation\end{tabular}  & SS         & Manipulate spreadsheets (~\cite{li2024SheetCopilot, chen2024SheetAgent})                                                                             & SpreadsheetBench~\cite{ma2024SpreadsheetBench}      \\ \hline
NL2SQL                                                               & DB         & Translate a NL question to a SQL query (~\cite{li2024CodeS, gao2024TexttoSQL})                                                                                          & Spider~\cite{yu2018Spider} \\ \hline
Data analysis                                                        & SS/DB      & \begin{tabular}[c]{@{}c@{}}Table data analysis pipeline, consists of\\ feature engineering, machine learning, etc. (~\cite{zhang2023DataCopilot, xu2024Lemur})\end{tabular} & DS-1000~\cite{lai2023DS1000}           \\ \hline
Table detection                                                      & DOC        & Locate tables in documents (~\cite{chen2023TableVLM})                                                                                                        & TableBank~\cite{li2020TableBank}         \\ \hline
Table extraction                                                     & DOC        & Extract and structuralize tables from documents (~\cite{zhao2024TabPedia, chen2023TableVLM})                                                                                   & PubTabNet~\cite{zhong2020Imagebasedb}         \\ 
\bottomrule
\end{tabular}
\end{table*}

Table~\ref{tab:table_tasks} summarizes table tasks that can be automated by LLMs. We also list the description of the table task, along with one or two related works, the types of tables addressed, and the datasets used.

Table QA and fact verification are the most traditional table tasks, which extract knowledge from tables to answer natural language (NL) questions. 
Table-to-text produces an NL text based on table data.
Data cleaning identifies and corrects errors in table data.
Column/Row/Cell population generates possible column/row/cell for a table.
Entity linking disambiguates specific entities mentioned, while column type annotation categorizes columns with types from knowledge bases.
These two tasks often utilize external knowledge bases.
Spreadsheet systems are originally designed for human users. Spreadsheet manipulation is a task that leverages AI to modify spreadsheets automatically, where AI accesses spreadsheet systems' APIs or formulas.
NL2SQL translates NL questions into SQL queries and can improve the efficiency of data analysts when writing SQL queries. This task has been extensively studied for years, and LLMs enhance accuracy in this field.
Data analysis consists of feature engineering, machine learning, etc.
Table detection identifies tables within documents, while table extraction converts them into table-native formats such as markdown, HTML, or spreadsheet.
The tasks mentioned above can broadly be categorized into table-related, spreadsheet-related, \\database-related, and document-related tasks. These tasks require AI models to directly understand table contents, write code to manipulate spreadsheets, write SQL to access databases, or extract table data from documents.

\subsection{Data Lifecycle and End-users' Perspective}
Researchers often concentrate on designing new \\methods to improve performance on benchmarks. 
However, end-users are primarily interested in how table-related AI systems can boost their productivity rather than benchmark results.
To meet end-users' needs, the industry focuses on developing products and tools for them.

\subsubsection{Data Lifecycle}

\begin{figure*}[ht]
    \centering
    \includegraphics[width=1\linewidth]{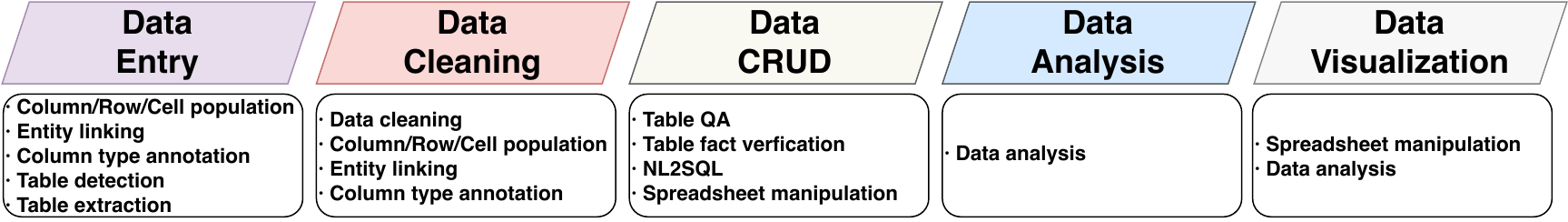}
    \caption{Table processing lifecycle with table tasks annotated.}
    \label{fig:data_lifecycle}
\end{figure*}

End-users' requirements vary based on their roles; common users typically need table querying and manipulation capabilities, while data engineers need data preparation and modeling tools. Different end users are at different stages of the data lifecycle. We divide the table data lifecycle into the following five stages: Data Entry, Data Cleaning, Data CRUD (Create, Read, Update, and Delete),  Data Analysis, and Data Visualization. Table~\ref{fig:data_lifecycle} shows the five stages in table data processing, with corresponding table tasks annotated below.

\textbf{Data Entry} consists of two parts, one is helping users to create the table structure, and the other is the precise entry of data by converting unstructured data formats into (semi-)structured tables. 
When creating a table, LLMs can help list the possible column headers. For instance, Google Sheets offers a feature that generates a new table with suggested column headers and example table data. Another application scenario involves converting tables from images or PDFs into formats natively suited for tables, facilitating subsequent stages of table processing. This feature requires AI systems capable of multimodal table understanding~\cite{zheng2024Multimodal, chen2023TableVLM}, and systems like ChatGPT-4o can now convert table images into structured formats.

\textbf{Data Cleaning} identifies and corrects errors, inaccuracies, missing values, and duplicates in a table dataset to improve its quality and reliability for further analysis~\cite{abedjan2016Detecting}. This stage needs to identify erroneous parts and impute errors or missing values, utilizing techniques like cell population or column type annotation.

\textbf{Data CRUD} includes the following tasks: table QA, table fact verification, NL2SQL, and spreadsheet manipulation.
This stage involves querying web table knowledge, transforming upstream data\-base tables into downstream tables in data warehouses or data lakes, or managing spreadsheet tables by calling the system's APIs or formulas. In this stage, AI systems usually enable individuals to process tables through NL questions or instructions.

\textbf{Data Analysis} includes feature engineering, outlier detection, machine learning, visualization, etc. Thus, it requires higher intelligence, as it involves understanding table data, having some domain kn\-owledge, and utilizing tools (e.g., SQL, Python, or VBA) to model tables and give insights.

\textbf{Data Visualization} is an essential step to improve the expressiveness of data. Different data types paired with distinctive chart types will show completely different expressiveness. Users expect that AI systems can automatically select the best chart type and graph descriptions.

\begin{figure}[ht]
    \centering
    \includegraphics[width=1\linewidth]{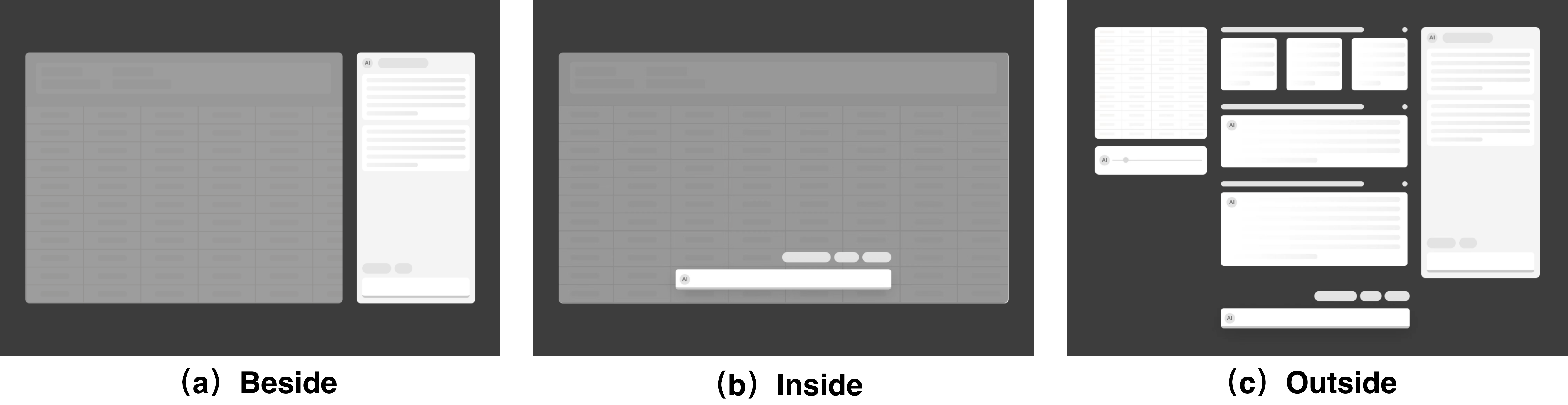}
    \caption{Three types of human interaction with table AI system: Beside, Inside, and Outside.}
    \label{fig:ui}
\end{figure}

\subsubsection{End-users' Experience}

As shown in Figure~\ref{fig:ui}, from the perspective of product design and usage, the interaction between end-users and AI systems can be categorized into three types: Beside, Inside, and Outside. 
``Beside" refers to adding a copilot beside the application.
``Inside" indicates that AI is the core component of the product.
``Outside" implies that the AI system orchestrates across different applications and tasks.
No matter the type, users tend to interact directly with AI through NL, and the quality of the user experience is strongly linked to the underlying table AI system~\cite{barke2023Grounded}.

\section{Table Data Representation}
\label{sec:representation}

When processing tables and table tasks, AI systems must first convert tables into machine-readable representations.
Modern neural networks, which are widely adopted in AI systems, need to encode text or images into numerical representations or embeddings, and then perform computations on these embeddings. Consequently, we must format table data accordingly and feed it into language or visual language models. 
This section focuses on the serialization of text tables and the processing of document tables.

\subsection{Text Representation}
\label{subsec:text_repr}

LLMs require prompts in a linear, sequential text format, contrasting with the 2-dimensional structure of tables with a defined schema. So, we should maintain their semantic integrity when converting tables into prompts. The prompt may contain the table content and the table schema. 

\textbf{Table Content} 
A straightforward yet common way is to linearize tables into markdown format, i.e., separating rows by new lines and inserting column separators (e.g., ``\textbar{}") between cells.
Several studies~\cite{sui2024Table}\cite{singha2023Tabular} evaluate different table serialization formats to assess whether LLMs accurately understand structured tables. 
They compare formats like CSV, markdown, JSON, HTML and pandas DataFrame.
The evaluation results suggest that HTML and NL with separators (markdown or CSV) are the two most effective options, which can be assumed that the training corpora include substantial code and web tables.
Spreadsheets may contain merged cells and hierarchical columns, meaning simple row-by-row and column-by-column serialization is insufficient. Tian et al.~\cite{tian2024SpreadsheetLLM} argue that homogeneous rows or columns in spreadsheets contribute little to understanding their layout and structure. Consequently, they introduce an anchor-based approach that pinpoints heterogeneous rows and columns as anchors. Subsequently, they utilize an inverted-index style encoding technique to convert cell locations, values, and schemas into a JSON dictionary format.

\textbf{Table Schema}
Some research~\cite{gao2024TexttoSQL,nan2023Enhancing} explores schema representation methods for NL2SQL tasks.
In NL2SQL tasks, the prompt should include the NL question, table schemas, instructions, etc. 
Table schemas can be represented in plain text or coded forms using \texttt{CREATE TABLE} statements.
Foreign keys can suggest the relationships among different relational tables.
Special prompt rules like ``{\it with no explanation}" force LLMs to provide clear and concise responses that align more closely with the standard answers in the benchmarks.
Results~\cite{gao2024TexttoSQL,nan2023Enhancing} show that information about foreign keys and the rules like ``{\it with no explanation}" instruction can benefit the NL2SQL task.

\textbf{Text Embedding} Like other text data, once text-based tables are serialized, the serialized table data will be embedded by LLMs.

\subsection{Visual and Layout Representation}
\label{subsec:visual_repr}

Web tables, spreadsheets, and document-embedded tables may have visual cues, such as color highlighting, which could potentially aid VLMs in identifying accurate table information.
Deng et al.~\cite{deng2024Tables} explore the capability of VLMs to comprehend image tables and assess the comparative performance of LLMs with text tables versus VLMs with image tables.
Their findings indicate that representing tables in image form can facilitate complex reasoning for VLMs.
To process these tables, AI systems require tools that convert visual cues into visual and layout embeddings.

\textbf{Preprocessing}
Some works like LayoutLM~\cite{xu2020LayoutLM, xu2021LayoutLMv2} leverage pre-built optical character recognition (OCR) tools and PDF parsers for preprocessing images and PDF files. Other research in multimodal table understanding involves converting tables into images.
For instance, Table-LLaVA~\cite{zheng2024Multimodal} converts HTML web tables into images to augment training data, enabling the model to understand tables within documents better.
Xia et al.~\cite{xia2024Vision} transform spreadsheets into images to explore the capabilities of VLMs in comprehending spreadsheets.

\textbf{Visual Embedding}
Visual embedding is the combination of image, position, and segment embeddings. To handle table images in documents, AI systems must encode images into features.
For example, LayoutLM and TableVLM~\cite{chen2023TableVLM} utilize Res\-Net~\cite{heDeepResidualLearning2016} to convert images into visual embeddings, whereas Table-LLaVA~\cite{zheng2024Multimodal} employs a Vision Transformer (ViT)~\cite{dos2021ViT}.

\textbf{Layout Embedding}
Layout embeddings capture the spatial layout information present in table images.
Both LayoutLM and TableVLM normalize and discretize coordinates to integer values within the range $[0, 1000]$, employing separate embedding layers for the x and y axes to represent the \\2-dimensional features distinctly.
\section{Table Training}
\label{sec:train_method}

In this section, we explore the training techniques of large models for table tasks. 
There are primarily two types of LLMs: large language models (LLMs) that accept only text input and visual language models (VLMs) that can process visual inputs. 
Given the differences in inputs, model architecture, and training techniques between these two types, we summarize table training techniques in Fig.~\ref{fig:table_llm_train} and will discuss each category individually.
We begin by reviewing the literature on small language models (SLMs) prior to the LLM era, where these studies train models with fewer than a billion parameters.
We will then delve into the details of LLMs and VLMs for tables.
\subsection{Pre-LLM Era}

Before the era of LLMs, many researchers were already employing language models to address table tasks.
Their works primarily focus on modifying model structures, devising encoding methods, and designing training objectives to tailor the models for table tasks.
For example, TaPas~\cite{herzig2020TaPas} extends BERT's~\cite{devlinBERTPretrainingDeep2019} model architecture and mask language modeling objective to pre-train and fine-tune with tables and related text segments.
TaBERT~\cite{yin2020tabert} encodes a subset of table content most relevant to the input utterance and employs a vertical attention mechanism.
TURL~\cite{deng2020TURL} encodes information of table components (e.g., caption, headers, and cells) into separate input embeddings and fuses them together.
TABBIE~\cite{iida2021tabbie} modifies the training objective to detect corrupted cells.
TaPEx~\cite{liu2022TAPEX} learns a synthetic corpus, which is obtained by automatically synthesizing executable SQL queries and executing these SQLs.
RESDSQL~\cite{li2023RESDSQL} injects the most relevant schema items into the model during training and ranks schema items during inference to find the optimal one.

To process image tables and tables within documents, researchers often adopt the encoder-decoder architecture. In this architecture, the encoder encodes visual information while the decoder generates textual output.
For instance, 
LayoutLM~\cite{xu2020LayoutLM, xu2021LayoutLMv2} integrates textual, visual, and layout embeddings into its BERT backbone.

However, the foundation models of these methods are relatively small. Some cannot adapt to various downstream tasks, and some require annotated data during fine-tuning.

\subsection{Table LLM Training}
\label{subsec:llm_training}

\tikzstyle{my-box}=[
    rectangle,
    draw=hidden-draw,
    rounded corners,
    text opacity=1,
    minimum height=1.5em,
    minimum width=5em,
    inner sep=2pt,
    align=center,
    fill opacity=.5,
    line width=0.8pt,
]
\tikzstyle{leaf}=[my-box, minimum height=1.5em,
    fill=hidden-pink!80, text=black, align=left,font=\normalsize,
    inner xsep=2pt,
    inner ysep=4pt,
    line width=0.8pt,
]

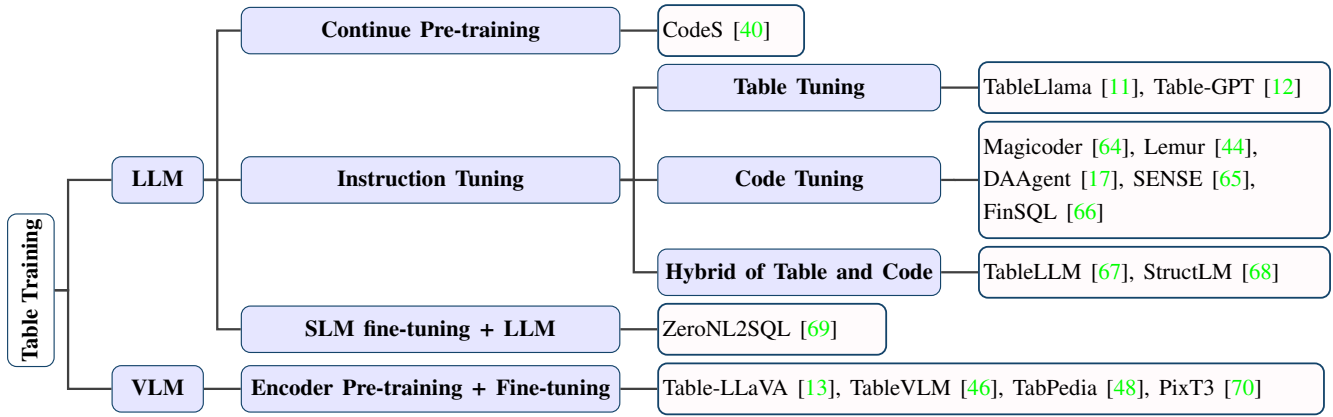
\begin{figure*}[t!]
    \centering
    \resizebox{\textwidth}{!}{
        \begin{forest}
            forked edges,
            for tree={
                grow=east,
                reversed=true,
                anchor=base west,
                parent anchor=east,
                child anchor=west,
                base=center,
                font=\large,
                rectangle,
                draw=hidden-draw,
                rounded corners,
                align=left,
                text centered,
                minimum width=3em,
                edge+={darkgray, line width=1pt},
                s sep=3pt,
                inner xsep=2pt,
                inner ysep=3pt,
                line width=0.5pt,
                ver/.style={rotate=90, child anchor=north, parent anchor=south, anchor=center},
            },
            where level=1{text width=3em,font=\normalsize,}{},
            where level=2{text width=13.5em,font=\normalsize,}{},
            where level=3{text width=10em,font=\normalsize,}{},
            [
                \textbf{Table Training}, ver
                [
                     \textbf{LLM},fill=blue!10
                    [
                        \textbf{Continue Pre-training},fill=blue!10
                        [
                         CodeS~\citep{li2024CodeS}, leaf, text width=5em
                        ]
                    ]
                    [
                        \textbf{Instruction Tuning}, fill=blue!10
                        [
                            \textbf{Table Tuning}, fill=blue!10
                            [
                        TableLlama~\citep{zhang2023TableLlama}{,}
                        Table-GPT~\citep{li2024TableGPT}, leaf, text width=12.5em
                            ]
                        ]
                        [
                            \textbf{Code Tuning}, fill=blue!10
                            [
                        Magicoder~\cite{wei2024Magicoder}{,}
                        Lemur~\cite{xu2024Lemur}{,}\\
                        DAAgent~\cite{hu2024InfiAgentDABench}{,}
                        SENSE~\cite{yang2024Synthesizing}{,} \\
                        FinSQL~\cite{zhang2024FinSQL}, leaf, text width=12.5em
                            ]
                        ]
                        [
                           \textbf{Hybrid of Table and Code}, fill=blue!10
                            [
                        TableLLM~\cite{zhang2024TableLLM}{,}
                        StructLM~\cite{zhuang2024StructLM}, leaf, text width=12.5em
                            ]
                        ]     
                   ]
                   \textbf{LLM},fill=blue!10
                    [
                        \textbf{SLM fine-tuning + LLM},fill=blue!10
                        [
                         ZeroNL2SQL~\citep{fan2024Combining}, leaf, text width=8em
                        ]
                    ]
                ]
                [
                    \textbf{VLM},fill=blue!10
                    [
                        \textbf{Encoder Pre-training + Fine-tuning},fill=blue!10
                        [
                         Table-LLaVA~\citep{zheng2024Multimodal}{,} 
                         TableVLM~\cite{chen2023TableVLM}{,} 
                         TabPedia~\cite{zhao2024TabPedia}{,} 
                         PixT3~\cite{alonso2024PixT3}, leaf, text width=24em
                        ]
                    ]
                ]
            ]
        \end{forest}
    }
    \caption{Summary of Table LLM training techniques.}
    \label{fig:table_llm_train}
    \vspace{0mm}
\end{figure*}


\subsubsection{What's New in Table LLM Training}

Methods such as instruction tuning and continued pre-training are widely utilized in the era of LLMs. Although these approaches have already been employed in the pre-LLM era, their training techniques differ from those used previously. 

\textbf{Instruction tuning} involves directing the model with specific instructions or guidance within the prompt. It enables the language model to follow the instructions given in the prompt.
As shown in Fig.~\ref{fig:table_llm_train}, for table processing tasks, there are three types of instruction tuning: table tuning, code tuning, and a hybrid of both table and code.


\begin{itemize}
    \item 
\textbf{Table Tuning} focuses on LLMs' understanding of tables so that LLMs can handle various table tasks such as table QA, table-to-text, entity linking, etc.
This type of research utilizes general-purpose foundation LLMs \\(e.g., Llama) and a substantial volume of table-related data for instruction tuning.  Table tuning examples include TableLlama\cite{zhang2023TableLlama} and \\Table-GPT\cite{li2024TableGPT}.

    \item 
\textbf{Code Tuning} addresses table processing tasks from a code generation perspective by generating code (e.g., SQL or Python) to manipulate table data.
Code tuning examples include Magicoder~\cite{wei2024Magicoder}, Lemur~\cite{xu2024Lemur} and DAAgent~\cite{hu2024InfiAgentDABench}.

    \item 
\textbf{Hybrid of Table and Code} research reveals that table instruction tuning based on code LLMs is more effective, i.e., code LLMs are tuned using table instruction datasets. For example,
StructLM~\cite{zhuang2024StructLM} and TableLLM~\cite{zhang2024TableLLM} tune code LLMs on table datasets.
\end{itemize}

Building Instruction Datasets is of great importance for high-quality instruction tuning. 
Manually crafting these datasets is highly time-consuming and labor-intensive, presenting a major challenge in constructing high-quality datasets. So research focuses on constructing datasets via automatic approaches like using templates to transform existing annotated datasets for instruction tuning or distilling data from more powerful LLMs. 
Therefore, researchers concentrate on constructing instruction datasets automatically, employing methods such as template-based transformation of existing annotated datasets, or data distillation from more powerful LLMs. 

\textbf{Continue Pre-training} takes an existing trained model, feeds new data, and adapts it for a specific task~\cite{parmar2024Reuse}. \textbf{SLMs + LLMs} is a paradigm to fine-tune an SLM to guide the LLM towards desired outputs where two types of models complement each other~\cite{li2023Guiding}.

\subsubsection{Table Tuning}

Table tuning, short for table instruction tuning, constructs instruction tuning datasets by leveraging multiple existing table-related datasets.
The instruction tuning dataset can be in the form of {\it (Instruction, Table, Output)}. 
Fig.~\ref{fig:table_it} is an example entry from TableInstruct, the instruction tuning dataset for TableLlama~\cite{zhang2023TableLlama}.
In this example, 
the {\it Instruction} specifies the task; the {\it Table} describes table content, table metadata, or task-specific content. 
The {\it Output} features natural language outputs like table QA answers, text from table-to-text transformations, result tables after manipulation, or a mix of text and tables.
This example is about fact verification and has a NL question. For other table tasks, the {\it Question} element may be optional.
When using web tables, titles or captions are included in the table content to provide context information to LLMs.

\begin{figure}[ht]
\begin{tcolorbox}[colback=white, colframe=black, title=An example entry from \texttt{TableInstruct}, fontupper=\footnotesize, fonttitle=\footnotesize]
\textbf{\#\#\# Instruction}: \\
This is a table fact verification task. The goal of this task is to distinguish whether the given statement is entailed or refuted by the given table. \\
\textbf{\#\#\# Table}: \\
\text{[TLE]} The table caption is about tony lema. [TAB] \textbar ~tournament \textbar ~wins \textbar ~ top - 5 \textbar ~top - 10 \textbar ~top - 25 \textbar ~events \textbar
~cuts made [SEP] \textbar ~masters tournament \textbar ~0 \textbar ~1 \textbar ~2 \textbar ~4 \textbar ~4 \textbar ~4 \textbar ~[SEP] ... \\
\textbf{\#\#\# Question}: \\
The statement is:\textless tony lema be in the top 5 for the master tournament, the us open, and the open championship\textgreater. Is it entailed or refuted by the table above? \\
\textbf{\#\#\# Output}: \\
Entailed.
\end{tcolorbox}
\caption{An example entry from \texttt{TableInstruct}, a table instruction tuning dataset.}
\label{fig:table_it}
\end{figure}

TableLlama emphasizes using more realistic data and uses the {\it template} approach to collect 14 existing datasets (e.g., WikiTableQuestions~\cite{pasupat2015wikitq} or Spider~\cite{yu2018Spider}) of 11 table tasks. 
Table-GPT employs the {\it synthesis-then-augment} approach. 
The synthesis-then-augment approach resembles the method used in computer vision, where images are randomly \\cropped or flipped to create variations.
Table-GPT designs 18 synthesis processes, ranging from table QA to row/column swapping. For example, the row/column swapping synthesis process swaps rows or columns, and the {\it Output} is the swapped table. 
In this way, the model can understand the order of rows/columns.
Additionally, Table-GPT implements augmentation strategies at the instruction level, table level, and output level to increase task and data diversity. For example, the instruction level augmentation uses powerful LLM to paraphrase the canonical human-written instruction into many different variants. These augmentation approaches can prevent the model from overfitting.
TableLlama and Table-GPT demonstrate that after table tuning on seen table tasks, LLMs could exhibit robust generalization capabilities and tackle unseen table tasks.

\subsubsection{Code Tuing} 

Tables can be manipulated through programming languages such as Python or SQL. Consequently, the code instruction tuning approach develops LLMs specializing in code generation.
Models tuned with table instructions typically generate table-related content directly.
Code LLMs first generate code that is subsequently executed in environments like Python interpreters or database engines.
Code LLMs are particularly adept at tasks like NL2SQL and data analysis. Several code LLMs achieve top rankings on the DS-1000~\cite{lai2023DS1000} and InfiAgent-DABench~\cite{hu2024InfiAgentDABench} leaderboards, which are benchmarks for data analysis code generation. Here, we list a few examples and discuss how they build instruction datasets.



WizardCoder~\cite{luo2024Wizardcoder} employs the ``Evol-Instruct" method, where ``Evol" denotes evolution, indicating the use of existing instruction data as a seed to prompt a more powerful LLM to generate new instructions. Fig.~\ref{fig:evol_ins} shows a sample prompt from Evol-Instruct for code. \texttt{Problem} refers to the current code instruction awaiting evolution and \texttt{Method} is evolution type. WizardCoder uses five heuristic evolution methods, and here in Fig.~\ref{fig:evol_ins}, we provide one example heuristic method.

\begin{figure}[ht]
\begin{tcolorbox}[colback=white, colframe=black, title=A sample prompt from \texttt{Evol-Instruct} for code, fontupper=\footnotesize, fonttitle=\footnotesize]
Please increase the difficulty of the given programming test question a bit. 
You can increase the difficulty using, but not limited to, the following methods: 
\\
\textbf{\#\#\# Method}:
\\ Provide a piece of erroneous code as a reference to increase misdirection.
\\
\textbf{\#\#\# Problem}:
\\ \{problem\}
\end{tcolorbox}
\caption{A sample prompt from \texttt{Evol-Instruct} for code.}
\label{fig:evol_ins}
\end{figure}

Magicoder~\cite{wei2024Magicoder} propose a process called ``OSS-Instruct". OSS-Instruct first collects open-sourced code snippets and lets a powerful LLM draw inspiration from the code snippets to produce realistic code instructions.
OSS-Instruct is orthogonal to existing data generation methods like Evol-Instruct. Thus, these two methods can be combined together.
Both WizardCoder and Magicoder utilize powerful LLMs to distill and generate additional data.
Lemur~\cite{xu2024Lemur} argues that code LLM should balance between the general-purpose ability and the code ability. The general-purpose ability is for reasoning and planning and can be learned by NL text. The code ability, which can be learned from code, ensures grounding in programming environments. So, the authors build a corpus with a 10:1 code-to-text ratio to ensure that the trained code LLMs have coding ability while maintaining performance in NL ability.
Lemur also evaluates whether the model performs effectively with agents that heavily depend on tool usage and environment feedback.
DAAgent~\cite{hu2024InfiAgentDABench} is a series of specialized agent models focused on data analysis. Their instruction tuning dataset is crafted by crawling CSVs from GitHub and generating data analysis keywords and questions by iteratively prompting GPT-4 given a specific crawled CSV file.
These studies show that enhancing models with code instructions can boost performance on table tasks, especially on data analysis tasks. 
When constructing code instruction\\datasets, these methods more or less distill data from powerful LLMs (e.g., GPT).

SENSE~\cite{yang2024Synthesizing} is an NL2SQL model that utilizes two types of synthetic training data: ``strong data" and ``weak data."
``Strong data" is distilled from powerful LLMs that provide more reliable responses, while ``weak data" is produced by inferior models that may result in errors during SQL execution.
The strong and weak data are then trained with Direct Preference Optimization (DPO)~\cite{rafailov2023Direct}, enabling the SENSE model to learn from both correct and incorrect samples.
FinSQL is also an NL2SQL solution. It creates a dataset specialized for the financial sector and employs a powerful LLM to augment the data, followed by parameter-efficient fine-tuning (PEFT)~\cite{lester2021Power, hu2022LoRA} of a LLM.

\subsubsection{Hybrid of Table and Code}

Currently, there are many open-source general-purpose models and code LLMs available.
One question that arises is which foundation model should be selected for further training specifically for table tasks.
StructLM~\cite{zhuang2024StructLM} performs an ablation study using code LLM, general LLM, and math LLM as foundational models, fine-tuning them on tabular datasets.
Studies~\cite{li2024Dawn, zhang2024TableLLM}, including StructLM reveal that using the code LLM as the foundation model achieves superior performance on table tasks.
Like table instruction tuning, studies of the hybrid type focus on how to construct table instruction datasets.

\subsubsection{Continue Pre-training}

Small-sized LLMs have lower deployment costs, but their code generation or reasoning abilities are inferior to those of large-sized LLMs.
CodeS~\cite{li2024CodeS} proposes continued pre-training~\cite{parmar2024Reuse} on small-sized LLMs to enhance their performance in NL2SQL tasks.
Specifically, CodeS feeds SQL-related, NL text, and NL-to-code data into the pre-trained StarCoder models~\cite{li2023starcoder}, thereby enhancing the models' capabilities in natural language processing, reasoning, and coding.

\subsubsection{SLMs + LLMs}
SLMs are easier to fine-tune to understand table schemas, whereas LLMs exhibit strong reasoning capabilities but may encounter ``hallucination"~\cite{ji2023Survey} problems owing to the lack of domain knowledge. ZeroNL2SQL~\cite{fan2024Combining} combines the strengths of both SLMs and LLMs to mitigate their respective weaknesses. It fine-tunes an Encoder-Decoder SLM responsible for generating SQL sketch candidates, and it employs an LLM to fill in missing parts in the SQL sketch, correct errors in the SQL query, and generate the final query.

\subsection{Table VLM Training}
\label{subsec:vlm_training}

Research in utilizing VLMs for table tasks can be divided into three types. The first type adheres to the conventional pattern recognition method~\cite{nassar2022TableFormer, xu2020LayoutLM}, which involves table detection and extraction tasks. A notable example is the TableVLM~\cite{chen2023TableVLM}. 
The second approach tackles various table tasks in an end-to-end manner. The example is Table-LLaVA~\cite{zheng2024Multimodal}.
The third type integrates features of the first two, enabling the detection, extraction, and end-to-end tasks such as question answering on image tables, with TabPedia~\cite{zhao2024TabPedia} serving as an example.
The third type is a hybrid of the first two, which can detect and extract tables and answer questions based on table images, exemplified by TabPedia~\cite{zhao2024TabPedia}.

\textbf{Pre-training + Fine-tuning} Most LLMs are designed with a decoder-only architecture. Similar to the network architecture of visual language models from the pre-LLM era, VLMs typically employ an encoder-decoder architecture, where a visual encoder converts visual data into embeddings while the decoder generates texts.
The encoder may utilize architectures such as ResNet or ViT, while the decoder typically consists of a pre-trained LLM.
Compared to the decoder, which has been thoroughly trained, the encoder has not yet mastered much visual information about tables. 
It is necessary to align the visual cues with the textual information. 
Therefore, the training process for a table VLM is usually divided into two phases: 1) Pre-training the visual encoder while freezing the parameters of the LLM decoder and 2) Fine-tuning or instruction tuning the entire model.
Both phases require a substantial amount of high-quality training data.
PixT3~\cite{alonso2024PixT3} is a multimodal table-to-text model that takes table-to-text tasks as table visual recognition tasks and generates texts, removing the need to process tables in text formats.

 

\section{Table Prompting}
\label{sec:prompt}

In this section, we explore the strategies for prompting large models to handle table tasks.
LLMs struggle with functionalities, such as complex reasoning, arithmetic calculation, factual lookups, and correcting erroneous decisions, all essential for table tasks. 
Thus, the key challenges include guiding the model towards complex reasoning, enabling it to reflect and revise rather than fast thinking, and utilizing external tools for executing Python or SQL code.
To address the issues above, researchers have been dedicated to developing LLM-powered agents.

\subsection{Common Workflow of LLM-powered Agents}

\begin{figure}[t]
    \centering
    \includegraphics[width=0.99\linewidth]{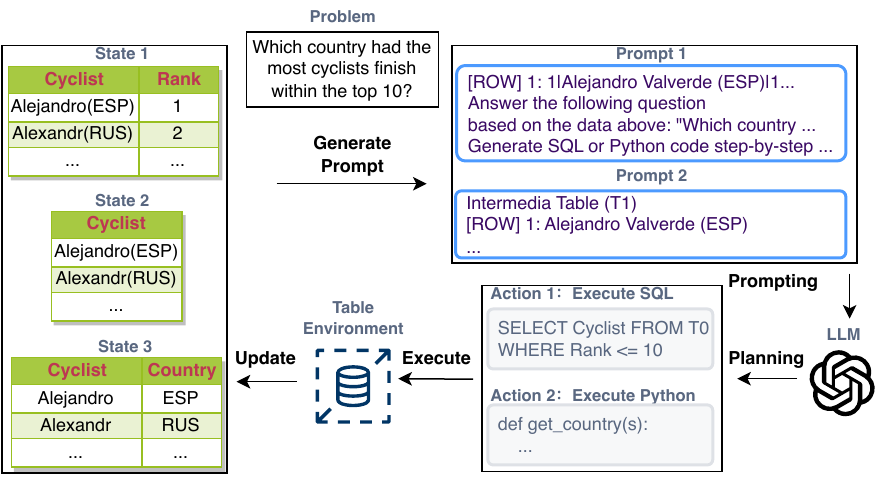}
    \caption{The iterative process of an LLM-powered agent system for table tasks.}
    \label{fig:agent}
    \vspace{-0.2in}
\end{figure}

Research such as Chain-of-Thought (CoT)~\cite{wei2022Chainofthought} and ReAct~\cite{yao2023ReAct} prompt LLMs iteratively, organizing the reasoning process into multiple intermediate \\steps.
Thus, LLMs address simpler subproblems step by step and progressively build a coherent response.
Agent systems are designed to follow this paradigm and typically include modules such as \textbf{memory}, \textbf{planning}, and \textbf{action}~\cite{wang2023AgentSurvey}.
The memory module stores state or observations from the environment or records past actions.
This information can be utilized for future planning.
The planning module chooses which action the agent needs to do in the current step, while the action module interacts with the environment and executes the action to get the outcomes.
As shown in Figure~\ref{fig:agent}, for table tasks, the agent first observes the table data and user intent. It generates prompts, decomposes complex tasks, plans actions, executes them on the table environment, and then updates the state or observation.
This iterative process is repeated until expectations are met.
Studies like SheetAgent~\cite{chen2024SheetAgent}, Chain-of-Table~\cite{wang2024chainoftable}, ReAcTable~\cite{zhang2024ReAcTable}, TAPERA~\cite{zhao2024TaPERA}, and $E^{5}$~\cite{zhang2024Zeroshot} follow this workflow.

The memory module enables the agent to accumulate experiences, self-evolve, and act with greater consistency, rationality, and effectiveness~\cite{wang2023AgentSurvey}. The memory of table agents stores planning history, that is, the outcomes of specific actions and table states, typically in a structured table format. In other aspects, the memory module of table agents is not significantly different from that of other LLM-based agents. 
This paper focuses on the planning and action modules of table agents.

\subsection{Planning}
\label{subsec:planning}

The planning module plans actions by prompting LLMs.
It must carefully handle two key aspects: 1) breaking down complex problems into smaller sub-problems, and 2) reflecting on and revising previous decisions.

\subsubsection{Formalizing Planning}

In table agents, the planning module interacts with the target tables using a ReAct approach, with feedback and reflection. SheetAgent~\cite{chen2024SheetAgent} provides a formal definition of planning for spreadsheet manipulation, and here, we further extend it to encompass more table tasks like table QA and NL2SQL. 
Typically, the input of the planning module usually consists of the task instruction $I$, table state $S_{t}$ at step $t$, user query or current (sub-)problem $Q$, planning history $H_{t-1}$. 
Utilizing this information, the planning module prompts LLMs and formulates an action $A_{t}$ for step $t$ by:

$$
A_{t} = P(A_{t} | I, S_{t}, Q, H_{t-1})
$$

The action, which we will discuss in Section~\ref{subsec:action}, is executed on the target table, yielding a new observation and the output denoted as $O_{t}$.
The table state and the planning history are then updated by $H_{t} = (H_{t-1}, O_{t}, A_{t})$.

\subsubsection{Complex Task Decomposition}

Inspired by the CoT and least-to-most~\cite{zhou2023Leasttomost} prompting methods, researchers instruct LLMs to decompose complex table tasks into simpler sub-tasks. 
DIN-SQL~\cite{pourreza2023Dinsql} proposes breaking down the NL\-2SQL task into subtasks. The process involves identifying the relevant tables and columns associated with the query and producing intermediate sub-quer\-ies. 
DEA-SQL~\cite{xie2024Decomposition} and TabSQLify~\cite{nahid2024TabSQLify} also follow the principle of decomposing tables into smaller, simplifying subtasks and reducing irrelevant information.
For web tables, Dater~\cite{ye2023Large} exploits LLMs as decomposers by breaking down huge evidence (a huge table) into sub-evidence (a small table) and decomposing a complex question into simpler sub-questions and getting SQLs.

\subsubsection{Reflection and Revision}

LLMs often generate answers ``without thinking", while agents can try different approaches, vote to select the best one, and even reflect on past actions, learn from mistakes, and refine them for future steps.

\textbf{Self-consistency and Voting}
The {\it self-consistency} ~\cite{wang2023Selfconsistency} decoding strategy plans multiple reasoning paths and chooses the most consistent answer. It can significantly enhance LLM's accuracy on complex tasks. 
This has also been demonstrated in table tasks, which are reported in papers of Dater~\cite{ye2023Large}, DAIL-SQL~\cite{gao2024TexttoSQL}, MCS-SQL~\cite{lee2024MCSSQL}, and \\ReAcTable~\cite{zhang2024ReAcTable} on table QA and NL2SQL tasks.
However, several studies~\cite{gao2024TexttoSQL,zhang2024ReAcTable} argue that while self-consistency and voting enhance accuracy, these methods are time-consuming, and the cost is higher, considering that prompting LLMs is not cheap.

\textbf{Revising}
In many table tasks, a mistake in one step's action can significantly impact the following analysis and processing. 
To handle this challenge, SheetCopilot~\cite{li2024SheetCopilot} and SheetAgent~\cite{chen2024SheetAgent} adopt a mechanism that reflects and refines past actions. 
SheetAgent features a Retriever component that retrieves high-quality code examples from their curated repository, which helps to prevent the generation of erroneous code.
SelfEvolve~\cite{jiang2023SelfEvolve} achieves decent results on the Python data analysis task by asking the LLM to perform debugging if the generated code cannot execute in the Python interpreter.
These techniques prevent error actions like incorrect APIs or wrong arguments.

\subsection{Action}
\label{subsec:action}

In table tasks, actions are intermediaries between LLMs and software tools like database engines, spreadsheet systems, or Python interpreters. 
Utilizing these external tools involves more than just converting NL text into software APIs. The agent system must ensure that the API calls are error-free and can address complex tasks.
These external tools extract data from tables and are similar to the Retrieval-Augmented Generation (RAG)~\cite{karpukhin2020Dense} concept.
Given the inherent characteristics of different table tasks, we will discuss how to define actions based on the specific tasks.

\subsubsection{Table QA and NL2SQL}

For tasks like table QA and NL2SQL, agent systems usually utilize Python or SQL to interact with tables.
In Binder~\cite{cheng2023Binding}, actions are categorized into two types: extended Python and SQL, which allow the injection of LLM as operators within standard Python and SQL. 
ReAcTable~\cite{zhang2024ReAcTable} incorporates three kinds of actions: (1) generating a SQL query, (2) generating Python code, and (3) directly answering the question.
It extends the ReAct framework with an {\it observation-action-reflection} loop specifically for table tasks. 
Within this iterative loop, it progressively refines the table data by generating and executing an SQL query to retrieve table knowledge.
If the existing table lacks the required information or if the answer cannot be directly \\queried via an SQL query, it generates and executes Python code to produce an intermediate table, thereby filling missing information into an intermediate table.

\subsubsection{Spreadsheet Manipulation and Data Analysis}

Spreadsheet manipulation and data analysis are \\highly flexible. First, users' expectations and requirements vary. Second, there is a wide range of operations and software APIs. Third, a complex task may contain multiple operations, leading to dynamic alterations in the table content.

SheetCopilot~\cite{li2024SheetCopilot}, an agent system for spreadsheets, models existing spreadsheet VBA APIs into atomic actions. 
An atomic action comprises the API name, argument lists, document string, and several usage examples. 
These atomic actions are not specific to one system and can be implemented on different spreadsheet systems.	
The authors design these atomic actions by crawling spreadsheet-related questions online, embedding and clustering them into categories, and abstracting them to VBA APIs by choosing the most representative ones.
SheetAgent~\cite{chen2024SheetAgent} finds that Python and SQL are more appropriate for spreadsheet manipulation than VBA, owing to the two programming languages being more aligned with the training data of existing LLMs. 
SheetAgent comprises two essential components for agent actions: a Planner and an Informer. The Planner generates Python code and employs a ReAct-style reasoning approach to manipulate the spreadsheet table. The Informer serves as an evidence provider, supplying subtask-specific SQL queries to assist the Planner in tackling complex tasks.
Data-Copilot~\cite{zhang2023DataCopilot} is for data science and visualization. Its actions are two-level: the interface is high-level pseudo-code descriptions, and the code is low-level executable.
When designing interfaces, Data-Copilot generates code for each user request. It then assesses whether there are similarities among these requests that can be merged or abstracted.

\subsubsection{Multiple Table Tasks}

Most prompting methods are for a single table task; here, we list several studies that address multiple tasks.
These frameworks share common characteristics: they first explore the data, including the data itself and its schema, and then plan and optimize actions.

StructGPT~\cite{jiang2023StructGPT} is capable of solving table QA, NL2SQL, and knowledge graph QA by developing three types of actions that target web tables, databases, and knowledge graphs.
These actions function as a reader, extracting knowledge from the data, which is then used to assist the LLM in its future planning.
The actions for web tables include extracting data or column names, whereas for databases, the actions involve extracting table data or metadata.
UniDM~\cite{qian2024UniDM} manages a variety of data manipulation tasks, such as data cleaning, on data lakes through a three-action process. 
The first action is extracting context information, including table metadata and related records associated with the task, utilizing this information as demonstrations or background knowledge. 
The second action involves converting this context information into NL texts suitable for LLMs to reason. The final action employs prompt engineering to formulate the desired prompt.
TAP4LLM~\cite{sui2023TAP4LLM} is a pre-processing toolbox to generate table prompts. Its action includes: 1) selecting appropriate rows and columns from the table (called {\it Table Sampling}), 2) integrating them with other table data (relevant external knowledge, metadata) (called {\it Table Augmentation}), and 3) serializing this information to fit the context length of the LLM.
ChatPipe~\cite{chen2024ChatPipe} is a system designed to facilitate effortless interaction between users and ChatGPT for table data analysis tasks.
Initially, users upload their dataset along with their query.
ChatPipe assists in analyzing the dataset and suggests various data analysis and feature engineering operations.
Given the numerous potential data operations, ChatPipe employs a reinforcement learning-based method, utilizing a Deep Q-Network model~\cite{fan2020Theoretical} to learn and recommend the most suitable operations.

\subsection{Other Techniques}

This subsection discusses other prompting techniques like few-shot examples of in-context learning~\cite{brown2020Language} and role-play.

\subsubsection{Few-shot Examples Selection}

In-context learning refers to the ability of models to learn from examples within the context of prompts, an ability that emerges in LLMs.
The key to in-context learning is how to select and organize the most helpful demonstrative examples into the prompt. 
For the NL2SQL task, selecting the most relevant example can be achieved by choosing example que\-ries more related to the target NL question or example SQL more similar to the potential SQL. 
Nan et al.~\cite{nan2023Enhancing} argue that diversity should be considered in addition to similarity.
As the context length is limited, organizing all the example information (i.e., NL question, schema, SQL) into the prompt ensures the quality of examples but sometimes may exceed the context length. 
DAIL-SQL~\cite{gao2024TexttoSQL} proposes a method that balances the quality and token quantity by removing examples' schemas, which are token-cost.

\subsubsection{Role-play}

LLMs can be assigned roles when prompting them. For instance, Zhao et al.~\cite{zhao2023Large} test a prompt like "Suppose you are an expert in statistical analysis.".
Tapilot-Crossing~\cite{li2024TapilotCrossing} utilizes a multi-agent environment to generate user intents and simulate use cases from real-world scenarios. Within this environment, each agent takes on a unique role, such as Administrator, Client, Data Scientist, or AI Chatbot, interacting with each other to mimic a realistic data analysis context.

\section{Resources}

In this section, we summarize open-source datasets, benchmarks, and software, as these artifacts can facilitate the community's progress.

\subsection{Datasets and Benchmarks}

\begin{table*}[t]
\centering
\caption{Datasets and benchmarks for table processing tasks.}
\label{tab:datasets}
\begin{tabular}{cccc}
\toprule
Dataset           & Table Task                                                                           & Data Sources                                                                           & Size                               \\ \hline
RobuT~\cite{zhao2023RobuT}             & Table QA                                                                             & \begin{tabular}[c]{@{}c@{}}WikiTQ~\cite{pasupat2015wikitq}, \\ WikiSQL~\cite{zhong2017wikisql}, \\ SQA~\cite{iyyer2017sqa}\end{tabular}         & 138,149 perturbed examples         \\ \hline
BIRD~\cite{li2023bird}              & NL2SQL                                                                               & \begin{tabular}[c]{@{}c@{}}kaggle.com, \\ CTU Prague~\cite{motl2024CTU} \\ Open tables\end{tabular},                                                        & 81 DBs, 12,751 NL2SQL pairs        \\ \hline
Dr.Spider~\cite{chang2023drspider}         & NL2SQL                                                                               & Spider~\cite{yu2018Spider}                                                                                 & 200 DBs, 15K perturbed examples    \\ \hline
ScienceBenchmark~\cite{zhang2023ScienceBenchmark}         & NL2SQL                                                                               & Human + AI Augmented                                                                                 & 3 DBs, 6k NL2SQL pairs    \\ \hline
SheetCopilot~\cite{li2024SheetCopilot}      & Spreadsheet manipulation                                                             & superuser.com                                                                          & 28 SSs, 13k QA pairs               \\ \hline
SpreadsheetBench~\cite{ma2024SpreadsheetBench}  & Spreadsheet manipulation                                                             & \begin{tabular}[c]{@{}c@{}}4 Excel online forums, \\ e.g., excelfourm.com\end{tabular} & 912 instructions, 2,729 test cases \\ \hline
DS-1000~\cite{lai2023DS1000}           & Data analysis                                                                        & stackoverflow.com                                                                          & 451 problems                       \\ \hline
InfiAgent-DABench~\cite{hu2024InfiAgentDABench} & Data analysis                                                                        & github.com                                                                                 & 631 CSVs, 5131 samples             \\ \hline
Tapilot-Crossing~\cite{li2024TapilotCrossing}  & Data analysis                                                                        & kaggle.com                                                                               & 1176 user intents                  \\ \hline
AnaMeta~\cite{he2023anameta}           & \begin{tabular}[c]{@{}c@{}}Entity linking, \\ Column type annotation\end{tabular}    & \begin{tabular}[c]{@{}c@{}}public Web, \\ TURL~\cite{deng2020TURL}, \\ SemTab~\cite{jimenez-ruiz2020SemTab}\end{tabular}                                                                & 467k WT/SSs                        \\ \hline
GitTables~\cite{hulsebos2023gittables}         & \begin{tabular}[c]{@{}c@{}}Column type annotation, \\ Column population\end{tabular} & github.com                                                                                 & 1M CSVs                            \\ \hline
SchemaPile~\cite{dohmen2024SchemaPile}         & \begin{tabular}[c]{@{}c@{}}Column type annotation, \\ Column population \\ NL2SQL \\ Data analysis\end{tabular} & github.com                                                                                 & \begin{tabular}[c]{@{}c@{}}221k DB schemas \\ 1.7M table definitions\end{tabular}                               \\ \hline
ComplexTable~\cite{chen2023TableVLM}      & \begin{tabular}[c]{@{}c@{}}Table detection,\\ Table extraction\end{tabular}          & synthetically generated                                                                & 1M tables (png, HTML)           \\ \bottomrule
\end{tabular}
\end{table*}

Traditional benchmarks, such as WikiTQ~\cite{pasupat2015wikitq}, WikiSQL~\cite{zhong2017wikisql}, and Spider~\cite{yu2018Spider}, are already widely used and studied. We will not elaborate on those here but rather focus on new benchmarks. 
Table~\ref{tab:datasets} presents recently proposed datasets and benchmarks, along with their data sources, sizes.
We summarize the features of these new benchmarks as follows:

\begin{itemize}[leftmargin=*]
\item \textbf{Robustness} For most table tasks (such as Table QA), swapping rows and columns, or replacing column names with synonyms or abbreviations, should not affect the final results. For a large table (that cannot fit into the LLM's context), the placement of the wanted cell, whether at the beginning, end, or middle of the table, should not affect the query result.
To evaluate the robustness, Dr. Spider\cite{chang2023drspider} and RobuT\cite{zhao2023RobuT} are proposed.
RobuT reveals that the performance of all table methods degrades when perturbations are introduced, yet close-source LLMs (e.g., GPT) exhibit greater robustness. 

\item \textbf{Human Involved Labeling} 
Some new datasets, which target data analysis, require extensive manual annotation.
DS-1000~\cite{lai2023DS1000}, InfiAgent-\\DABench~\cite{hu2024InfiAgentDABench}, Tapilot-Crossing~\cite{li2024TapilotCrossing} are designed to assess data analysis tasks. They gather data from the internet and annotate it either automatically or semi-automatically.
DS-1000 collects questions from StackOverflow, assesses their usefulness manually, and curates to form the benchmark. 
The authors manually adapt the original questions by providing input and output context into test cases and rewriting problems to prevent LLMs from learning and memorizing the data.
InfiAgent-DABench invites human experts to evaluate the dataset quality and compare human-made and GPT-4 generated data analysis questions via multiple metrics.
Tapilot-Crossing aims to build a benchmark for real-world data analysis.
An issue that cannot be overlooked is the quality of these datasets.
Wretblad et al. conduct a thorough analysis of the NL2SQL dataset BIRD~\cite{li2023bird}, discovering inaccuracies in some of the gold SQLs and noise within certain NL queries~\cite{wretblad2024Understanding}.
The creators of the BIRD dataset introduce this noise during the dataset creation process.
After correcting these errors, Wretblad et al. find that complex prompting methods (e.g., DIN-SQL) might be less effective than simple zero-shot prompting.
This research reveals the potential unreliability of table datasets such as BIRD, considering they are generated through human labeling, a process prone to introducing noise and errors.

\item \textbf{Real-world Workload}
Most datasets are derived from the web or synthesized using templates. These publicly available online data are usually simple, as some of them are just tutorials for beginners. In contrast, tables in real-world scenarios are far more complex than these. 
ScienceBenchmark~\cite{zhang2023ScienceBenchmark} introduces a real-world benchmark developed in collaboration with SQL experts and researchers specializing in policy-making, astrophysics, and cancer research.
Given the scarcity of real-world data in these domain-specific data\-bases, ScienceBenchmark employs a data augmentation strategy that starts with hundreds of human-labeled NL2SQL pairs to create thousands more data points.
SpreadsheetBench~\cite{ma2024SpreadsheetBench} proposes a benchmark aimed at real-world scenarios, where the authors meticulously analyzed user questions from four Excel forums. They argue that their benchmark could assess the performance of handling complex user instructions.

\item \textbf{Larger Scale} AnaMeta~\cite{he2023anameta} is a large-scale table metadata dataset. GitTables~\cite{hulsebos2023gittables} downloads millions of CSV tables from GitHub and aligns them with knowledge bases. 
SchemaPile~\cite{dohmen2024SchemaPile} is a corpus with 221k database schemas and 1.7 million table definitions. 
These datasets can help evaluate whether the AI system could understand the table schema and benchmark tasks like column type annotation. They can also be fed into neural models as training data for solving various downstream tasks.
ComplexTable~\cite{chen2023TableVLM} contains more than 1 million image tables featuring complex structures and can be utilized for visual tasks related to tables.
\end{itemize}

\subsection{Open-source Software}

The academia and industry have developed numerous open-source software; here, we select a few to discuss. LlamaIndex~\cite{llamaindex}  is a popular and versatile RAG framework that, while supporting some table tasks, is not specialized in this area.
Both DB-GPT~\cite{xue2024dbgpt} and Vanna~\cite{vanna} are AI tools designed for database interactions. They allow users to train models or utilize their built-in prompting features.
PandasAI~\cite{pandasai} enables users to clean, query, and visualize pandas DataFrames using NL questions.
Most of these tools are designed for NL2SQL and table QA tasks, which enable users to query tables with NL texts. This trend underscores the demand among general users to utilize NL to enhance the efficiency of table processing.
RetClean~\cite{ahmad2023RetClean} is a tool that leverages LLMs for data cleaning operations, such as missing value imputation.
\section{Analysis and Discussion}

\begin{table*}[ht]
\centering
\caption{A comparative analysis of various methodologies was conducted using four benchmarks in table QA and NL2SQL tasks. The experimental setup and performance metrics are referenced from Zhang et al.~\cite{zhang2024TableLLM}. The number of parameters and training tokens are derived from the respective papers detailing each method.}
\label{tab:compare_qa_sql}
\begin{tabular}{@{}c|c|c|c|cccc|c@{}}
\toprule
\multirow{2}{*}{Type}                                                      & \multirow{2}{*}{Method} & \multirow{2}{*}{\begin{tabular}[c]{@{}c@{}}\# of\\ Parameters\end{tabular}} & \multirow{2}{*}{\begin{tabular}[c]{@{}c@{}}\# of Tokens\\ to Train\end{tabular}} & \multicolumn{4}{c|}{Benchmarks}    & \multirow{2}{*}{\begin{tabular}[c]{@{}c@{}}Avg. \#\\ of Infers\end{tabular}} \\ \cmidrule(lr){5-8}
                                                                           &                         &                                                                             &                                                                                  & WikiTQ & FeTaQA & WikiSQL & Spider &                                                                                  \\ \midrule
\multirow{2}{*}{SLM Training}                                              & TaPEX                   & 0.14B                                                                       & -                                                                                & 38.55  & -      & 83.90   & 15.04  & 1                                                                                \\
                                                                           & TaPas                   & 0.11B                                                                       & -                                                                                & 31.60  & -      & 74.20   & 23.05  & 1                                                                                \\ \midrule
\multirow{2}{*}{LLM Training}                                              & TableLlama              & 7B                                                                          & 3.3B                                                                             & 48.82  & 67.73  & 43.70   & -      & 1                                                                                \\
                                                                           & TableLLM                & 13B                                                                         & 1.1B                                                                             & 62.40  & 74.50  & \textbf{90.70}   & 83.40  & 1                                                                                \\ \midrule
\multirow{3}{*}{Prompting}                                                 & CodeLlama               & 13B                                                                         & -                                                                                & 43,44  & 57.24  & 38.30   & 21.88  & 1                                                                                \\
                                                                           & GPT-3.5                 & -                                                                           & -                                                                                & 58.45  & 71.18  & 81.70   & 67.38  & 1                                                                                \\
                                                                           & GPT-4                   & -                                                                           & -                                                                                & \textbf{74.09}  & \textbf{78.35}  & 84.00   & 69.53  & 1                                                                                \\ \midrule
\multirow{4}{*}{\begin{tabular}[c]{@{}c@{}}Prompting\\ Agent\end{tabular}} & StructGPT(GPT-3.5)      & -                                                                           & -                                                                                & 52.45  & 11.80  & 67.80   & \textbf{84.80}  & 3                                                                                \\
                                                                           & Binder(GPT-3.5)         & -                                                                           & -                                                                                & 61.61  & 12.77  & 78.60   & 52.55  & 50                                                                               \\
                                                                           & DATER(GPT-3.5)          & -                                                                           & -                                                                                & 53.40  & 18.26  & 58.20   & 26.52  & 100                                                                              \\
                                                                           \bottomrule
\end{tabular}
\end{table*}

In order to demonstrate the accuracy and costs of various methods to the readers, this section presents a comparative analysis and discusses the advantages and disadvantages of different approaches. 
\\Specifically, utilizing data from several papers~\cite{zhang2024TableLLM, ma2024SpreadsheetBench, li2024Dawn, xu2024Lemur, hu2024InfiAgentDABench, pang2024Uncovering}, we summarize both accuracy and cost for four table tasks: table QA, NL2SQL, spreadsheet manipulation, and data analysis.

\subsection{Discussion on LLM Training}
The advantages of training-based methods are that they offer great control, allowing enterprises to conduct private training and deployment without data leakage to third-party model service providers.
However, two issues can't be ignored: cost and accuracy. 

\textbf{Cost} 
The expense of pre-training or fine-tuning a large model is substantial. 
Fine-tuning a 7B model typically requires eight 80GB GPUs, and the number of training tokens dictates the total time needed for training. The training speed, measured in tokens per second, is affected by both the hardware and the software and is reported in several technical reports, such as Llama~\cite{touvron2023LLaMA}. Table~\ref{tab:compare_qa_sql} lists the number of parameters and training tokens for two LLM training approaches, thereby providing readers with an understanding of the training costs.
The number of parameters can also be used to estimate the minimal costs for private deployment. For instance, loading a 7B model with float16 format requires at least 14GB of memory, while serving with concurrent requests necessitates additional memory for the transformer architecture's key-value (KV) cache~\cite{kwon2023Efficient}.

\textbf{Accuracy} 
As indicated in Table~\ref{tab:compare_qa_sql}, across four benchmarks, the LLM training approach excels in only one, whereas prompting a robust LLM (GPT-4) is superior.
The 7B TableLlama can't outperform task-specific fine-tuning models on specific tasks (see the WiKiSQL column in Table~\ref{tab:compare_qa_sql}).
Another example is the NL2SQL task, where Li et al.~\cite{li2024Dawn} perform a systematic evaluation with various methods, comparing LLM-based and SLM-based solutions. Results show that there is no clear winner between LLM and SLM solutions on different metrics and domains. 
On data analysis tasks, instruction tuned models like Lemur~\cite{xu2024Lemur} and DAAgent~\cite{hu2024InfiAgentDABench} cannot surpass GPT-4.
Table-GPT continues to train on GPT-3.5 and achieves better results on all table-related tasks than GPT-3.5 and ChatGPT. However, the cost of training is \\prohibitively high for ordinary enterprise users who wish to deploy privately.
Regarding training data, the cost of manual annotation is also high. Although synthesis is a cheap choice, the data quality is another concern. A more prevalent approach is using GPT as a teacher model for data distillation. Evidently, GPT is the performance ceiling.

\subsection{Discussion on LLM Prompting}

The LLM-powered agent method combines the \\power of LLMs and the flexibility of external tools. On many table tasks, prompting strong LLMs like GPT still performs the best.

\textbf{Limited Transferability}
However, these appro\-aches often necessitate hard-coded prompt templates, which are long strings of instructions and demonstrations manually crafted through trial and error.
A given string prompt might not generalize well to other pipelines, LLMs, domains, or data inputs.
Thus, it has limited transferability.

\textbf{Cost}
As shown in the ``Avg. \# of Infers" column of Table~\ref{tab:compare_qa_sql}, agents must prompt LLMs repeatedly to fulfill users' expectations.
Ma et al.~\cite{ma2024SpreadsheetBench} conduct experiments on agents for spreadsheet manipulation, and the most effective agent solution is the one that utilizes the ReAct framework~\cite{yao2023ReAct}, executes code within the execution environment and offers feedback to LLMs upon failure. 
Deploying multiple rounds of ReAct prompting with execution feedback takes a substantial time.
Consequently, all these factors increase the total time and financial costs.

\textbf{Privacy Issue}
Most prompting methods involve requesting third-party model service providers, such as OpenAI, which could lead to data leakage, a situation many enterprises are unacceptable.
As open-source models advance, enterprises can also deploy open-source models or their own fine-tuned versions, thus gradually mitigating this issue.

\textbf{Structure Understanding}
Foundation models, without fine-tuning for tables, still have difficulty understanding table structure and hierarchy.
Pang et al. ~\cite{pang2024Uncovering} create a benchmark named TIS to assess how effectively LLMs seek information from tables.
They discover that LLMs struggle with tables that have complex hierarchies and exhibit a poor understanding of table structures, such as locating a specific regin in a two-dimensional table.
Notably, most LLMs perform at nearly random accuracy levels (around 50\%) in the table structure understanding task, while GPT-4 achieve 66.1\%.
\section{Challenges and Future Directions}

In this section, we outline some challenges and considerations for future research.

\subsection{Diverse User Input When Serving} 
User inputs refer not only to users' NL queries but also to the table schema and content.

\textbf{User Query} In real-world applications, user NL queries are frequently ambiguous.
For example, when users are unfamiliar with the table they want to query, they often pose general requests without a clear objective, such as ``help me analyze this table".
Users might also pose questions unrelated to the table within a table AI system.
Even when users have some knowledge of the table schema and content, they may struggle to formulate their query accurately, leading to ambiguity in the query sentences.

\textbf{Table Schema and Content}
In practice, table schemas and content are highly diverse and often proprietary. 
For instance, in a domain or industry with which the model has limited knowledge, it faces challenges in effectively transferring into the field.
Existing table training datasets are relatively simple, either consisting of simple table structures scraped from the web or synthesized based on powerful LLMs.
Constructing a complex and diverse training dataset is quite costly.
Ensuring that the training data covers real-world business scenarios poses a significant challenge.

Future table LLMs should adapt quickly and chea\-ply to real-world business needs.
Research directions include synthesizing high-quality training data that reflects the diverse needs of specific domains by cost-effective methods. 

\subsection{Slow and Deep Thinking}
Daniel Kahneman has revealed that slow and deep thinking is the underlying mechanism of the human brain for processing complex problems~\cite{kahneman2012Thinking}.
The chain-of-thought approach is considered to guide LLMs in solving complex reasoning problems, and the recent OpenAI’s o1 shows that LLMs can achieve substantial enhancements in performance by allowing them to generate extended internal chains-of-thought.
Additionally, this internal chain-of-thought can facilitate both the model training processes and the pure model prompting methods. 

\textbf{Training}
Typically, there is a considerable gap between the user instructions and the ultimate answer. Therefore, a chain-of-thought can serve as an internal mechanism to explain the process of deriving the final answer, potentially mitigating the training difficulty faced by LLMs. However, acquiring an accurate chain-of-thought is challenging. It can be prohibitively expensive when achieved through human annotation and difficult to ensure accuracy when automatically generated by LLMs themselves, particularly in tasks involving complex logical reasoning over tables. Exploring cost-effective methods to guarantee an accurate chain-of-thought process is a valuable area of research.

\textbf{Prompting}
In existing prompting methods on table tasks, researchers develop their own chain-of-thought workflows, which heavily rely on hard-coded prompt templates to solve specific table tasks.
This results in methods with weak transferability.
Given that OpenAI's o1 possesses inherent chain-of-thought abilities, prompting and agent methods must reconsider how to construct their workflows.
On the other hand, the chain-of-thought approach requires numerous inference iterations, which is time-consuming; future table prompting methods should balance inference times with accuracy.

\section{Conclusion}

This survey is the first comprehensive investigation into Large Language Models (LLMs) for table processing across various tasks, encompassing table QA, spreadsheet manipulation, data analysis, etc. 
We provide a summary and categorization of table tasks from both academic and end-user perspectives.
We explore the popular and essential techniques for table processing, including data representation, training, and prompting.
We collect and discuss resources like open-source datasets, benchmarks, and software.
Beyond reviewing existing work, we also identify several challenges within this domain that could inform and guide future research directions.

\bibliographystyle{fcs}
\bibliography{ref}

\end{document}